\documentclass[11pt,a4paper]{article}
\usepackage[hyperref]{naaclhlt2019}
\usepackage{MnSymbol}
\usepackage{times}
\usepackage{latexsym}
\usepackage{graphicx}
\usepackage{subfig}
\usepackage{url}
\usepackage{multirow}
\usepackage{wrapfig}
\usepackage[utf8]{inputenc}
\usepackage{wasysym}
\aclfinalcopy 
\def\aclpaperid{16} 

\title{Tracing Antisemitic Language Through Diachronic Embedding Projections: France 1789-1914}
\author{Rocco Tripodi\qquad Massimo Warglien\qquad Simon Levis Sullam\qquad Deborah Paci\\Ca' Foscari University of Venice\\\footnotesize\texttt{\{rocco.tripodi, warglien, levissmn, deborah.paci\}@unive.it}}
\normalfont
\date{}
\begin{document}
\maketitle
\begin{abstract}
We investigate some aspects of the history of antisemitism in France, one of the cradles of modern antisemitism, using diachronic word embeddings.  We constructed a large corpus of French books and  periodicals issues that contain a keyword related to Jews and performed a diachronic word embedding over the 1789-1914 period. We studied the changes over time in the semantic spaces of 4 target words and performed embedding projections over 6 streams of antisemitic discourse. This allowed us to track the evolution of antisemitic bias in the religious, economic, socio-politic, racial, ethic and conspiratorial domains.  Projections  show a trend of growing antisemitism, especially in the years starting in the mid-80s and culminating in the Dreyfus affair.  Our analysis also allows us to highlight the peculiar adverse bias towards Judaism in the broader context of other religions. 
\end{abstract}
\section{Introduction}
Word embeddings are widely used in many Natural Language Processing (NLP) tasks. They provide a machine-interpretable representation of lexical features. Their effectiveness in representing words semantics consists essentially in the ability of learning association patterns in the training dataset. For this reason the learned representations contain human-like biases \cite{caliskan2017semantics}. These biases can be detected easily and can be related to gender, ethic or racial aspects \cite{garg2018word,voigt2017language}.

Since the use of word embedding is ubiquitous in many commercial products such as search engines and machine translators, the research community has introduced different techniques to debias them \cite{bolukbasi2016man,zhao2018learning}, especially under the gender dimension. Despite these efforts debiasing word embeddings seems to be harder than expected. In fact, while \citet{bolukbasi2016man} and \citet{zhao2018learning} demonstrated that it is possible to debias specific gendered-words, even after the debiasing procedure, the geometry of the embeddings remains almost the same with respect to non gendered-words \cite{gonen2019lipstick}, preserving their original bias.

In this work, we turn these biases to the historian's advantage and shed light on some aspects of the history of antisemitism in France during the so called \emph{long XIX century}, between the French Revolution and the First Word War, using diachronic word embedding. This technique allows to capture diachronic conceptual changes and to analyse stereotyped historical biases. We tracked how historical events and publications influenced the construction of the collective imaginary related to the Jewish question. 

We assume that words do not have a fixed meanings. They can be used in different contexts to evoke a great variety of meanings using different connotational nuances. These multiple meanings are acquired (or lost) over time in correspondence to specific socio-political events. For example, one of the meanings of the word \emph{usurier} (i. e.; \emph{money lender}), as reported by the \emph{French Historical Dictionary}, refers to: \emph{the financial activities of the Jews} [who since the Middle Ages were], \emph{the only ones authorised to lend on pawns} \cite{DHLF2010}. This association derives from the fact that especially between the XVI and the XIX century, this word acquired a negative connotation, nurtured by anti-Jewish prejudice and stereotyping developing from the idea of an illegitimate interest attached to this activity. This image, as the above mentioned definition explains, was also fixed in the collective imaginary by Shylock, the Jewish protagonist in Shakespeare's \emph{Merchant of Venice} (1598).

In this work, we trace the conceptual changes of words related to the Jewish question. We collected a large corpus for this purpose, composed of thousands of books and newspapers published in France between 1789 and 1914. We used diachronic word embedding to represent the data, measures of local changes in the semantic space of different words, and embedding projections to quantify biases in different semantic spheres. The measurement of local changes is particularly suited for our study because we do not want to identify new meanings in the words related to the Jewish question, instead we want to trace how the context of their use changed and how these changes affected the representation of Jews at the time of the rise of modern antisemitism. Measuring biases over time is particularly interesting because it allows to connect them with antisemitic streams as identified by historians in the field \cite{wilson1982ideology} and operationalised by us.
\section{Related Work}
Models for capturing diachronic conceptual changes are associated with the distributional hypothesis \cite{harris1954distributional,firth1957synopsis,weaver1955translation}: the semantics of a word is defined by the context in which it is used. Following this assumption, different models have been presented, based on co-occurrence vectors \cite{sagi2009semantic,gulordava2011distributional,basile2016diachronic} or word embeddings \cite{kim2014temporal,kulkarni2015statistically,hamilton2016cultural}.

These works are brought together by the idea of analysing the contexts in which a word occurs and have culminated in the measures of \emph{semantic shift} and \emph{cultural drift}, proposed by \cite{hamilton2016cultural} and \emph{the law prototipicality} proposed by \citet{dubossarsky2015bottom}. Semantic shifts are regular linguistic processes such as semantic widening (e.g., \emph{dog}, that in Middle English was used to refer to dogs of a particular breed) \cite{bloomfield1933language}. This measure was used to derive two laws of semantic change: the \emph{law of conformity}: semantic change scales with a negative power of word frequency; and the \emph{law of innovation}: polysemous words have significantly higher rates of semantic change \cite{hamilton2016diachronic}. Cultural drifts involve local changes to a lexical form's use (e.g.: the changes in the meaning of the word \emph{cell}: \emph{prison cell} $\rightarrow$ \emph{cell phone}) \cite{hamilton2016cultural}. The \emph{law of prototipicality} was introduced by \citet{dubossarsky2015bottom}: it states that  prototypical words, words that are near to the centroid of a cluster in a semantic space, change slower than words that are in a peripheral position. The laws of \emph{conformity, innovation and prototipicality} have been questioned by \citet{dubossarsky2017outta}, who used controlled conditions to test them.

Different works that tried to measure, directly or indirectly, cultural drifts have been proposed recently. \citet{garg2018word} analysed gender and ethnic stereotypes in the United States during the 20th and 21st centuries, using word embeddings trained on the Google Books and Corpus of Historical American English (COHA) corpora. \citet{kozlowski2018geometry} used diachronic word embeddings to conduct macro-cultural investigation of social stereotypes. \citet{kutuzov2017temporal} attempted to model the dynamics of wordwide armed conflicts using word embeddings trained on a news corpus. \citet{zhao2017men} analyzed the amplification effect that learning models present on the gender dimension when trained on biased data.
%
%
\section{Motivations and historical background}
We have looked at linguistics representation of Jews in 19th century France, which was one of the cradles of modern antisemitism in Europe, i.e. of the mostly secularized and racial transformation of the centuries-old Christian prejudice against the Jews \cite{katz1980prejudice}.

Since the entry and gradual integration of the Jews in French society after the Revolution of 1789, the appearance of anti-Jewish texts, the rise of public controversies, or the burst of cases and scandals in which Jews were supposedly involved marked the emergence and spread of the Jewish question on the French scene, in what have been called \emph{antisemitic moments} or \emph{episodes} \cite{birnbaum2011anti}. Especially during the Third Republic, beginning in 1870, references to Jews entered the French public discourse in relation to a supposed growing influence of the Jews on political and economic affairs, the rise of anticlericalism in the face of Catholic France (for which Jews were considered responsible), the accusation of an alliance between Jews and Freemasonry.

This process reached its climax with the Dreyfus affair (1894), the unfounded accusation against a French army officer to have sold intelligence information to the German enemy (Dreyfus would be exonerated in 1906): the affair caused the heavy spread of antisemitic accusations and anti-Jewish movements of opinion \cite{wilson1982ideology}. Different streams of antisemitism ran accross French society throughout this time, together with a pro-Jewish reaction driven by the supporters of Dreyfus' innocence \cite{kalman2010rethinking}.

The publication in 1845 (republished in 1886) of Alfred Toussenel's \emph{Les Juifs rois de l'epoque} caused especially the rise of the so-called economic antisemitism, which accused the Jews of an increasing economic and financial influence, of which the Rothschilds were considered the protagonists and became a symbol. This accusation was later confirmed by the supposed Jewish role in the crash of the Catholic bank Union Générale (1882) and in the Panama corruption scandal (1892), together with the revival of nationalism tied to the Boulangist crisis \cite{sternhell1998droite}. These events generated a resurgence of antisemitism. In response to the growing secularization and anticlericalism, French Catholics revived an ancient tradition of religious antisemitism, marked in this time by the appearance of works such as Gougenot des Mousseaux's \emph{Le Juif, le judaïsme et la judaïsation des peuples chrétiens} (1869) and by the anti-Jewish campaigns of Catholic periodicals such as \emph{L'Univers} and \emph{La Croix}.

In 1886 the journalist Edouard Drumont published the hugely successful \emph{La France juive. Essai d'histoire contemporaine}, which described a French society under a greedy Jewish influence and control, painting in the style of a novelist (inspired by Balzac and by contemporary feuilletons or serialized novels) the contours of Jewish conspiracies. Although the subtitle of the work suggests an essay of contemporary history, on reading it is as if one is before an enormous cauldron of common place assumptions on Jews which includes Catholic, social, racial, economic, and conspiratorial anti-Semitism. The success of his work depended on the waves it made in the intellectual milieu of the era and its impact on the popular masses attracted by the synthesis of anti-Semitism of the right, of a church worried about laicisation, and anti-Semitism of the left, anti-capitalist and laical. This and other books by Drumont mixed Catholic, socio-political, ethic and conspirationist antisemitism, accusing Jews of all sorts of religious offenses, political machinations, moral perversions and secret plots \cite{kauffmann2008edouard}.

The combination of these streams of anti-Jewish accusations, prejudices and stereotypes would christallize - or reach its climax - in the Dreyfus affair. We suggest that the usage in print (books and periodicals) of the term \emph{juif} or other terms related to the Jewish question, all characterised by an adverse bias,  was especially connected to antisemitic tendencies. However, we should note that this vocabulary was also present at the time in Biblical and theological scholarship, art and art-historical publications, fictional and theatrical literature, medical treatises and the rising social sciences. References to Jews in the public discourse were therefore not necessarily mobilised in a political context with explicit antisemitics aims.
\begin{figure*}[!ht] 
  \subfloat[Periodicals distribution]{%
   \includegraphics[trim={1cm .25cm 1.5cm .15cm},clip,width=0.25\textwidth]{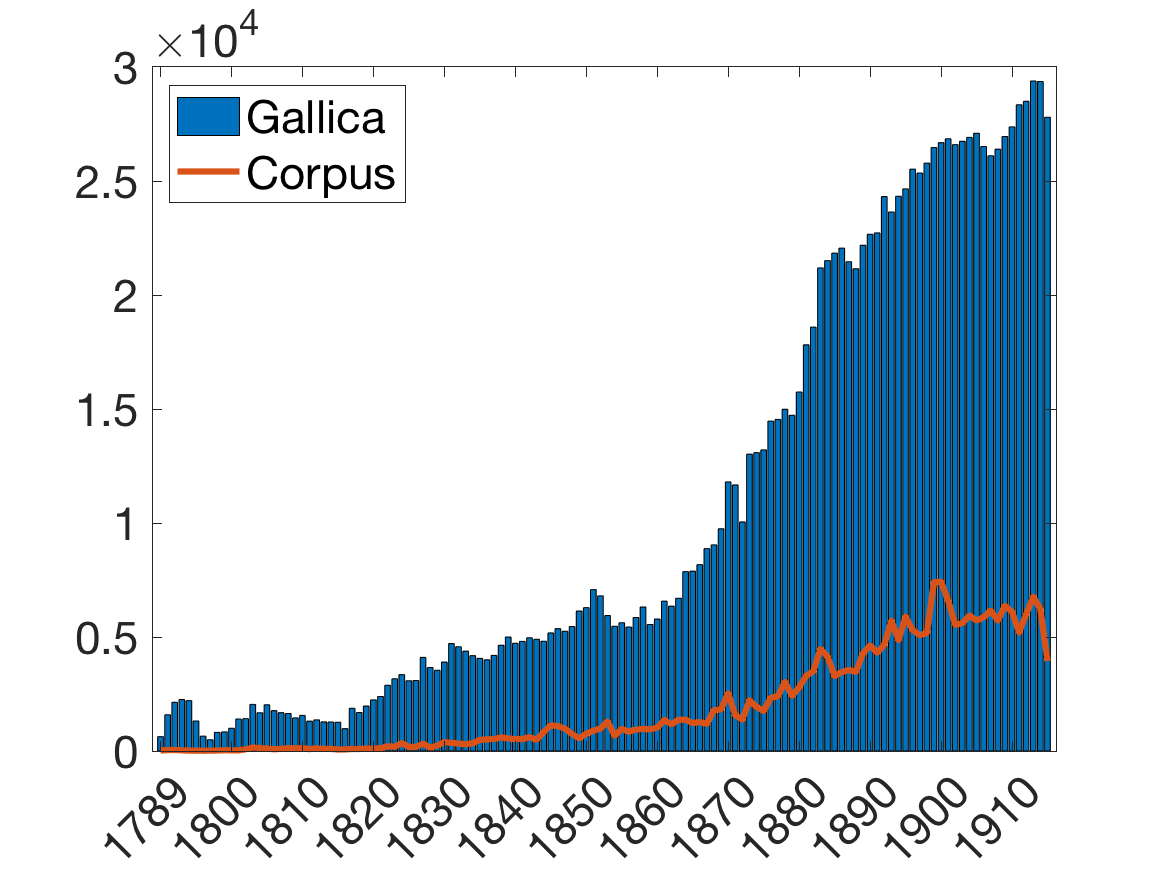}\label{fig:per_dist}}
  \hfill 
  \subfloat[Books distribution]{%
    \includegraphics[trim={1cm .25cm 1.5cm .15cm},clip,width=0.25\textwidth]{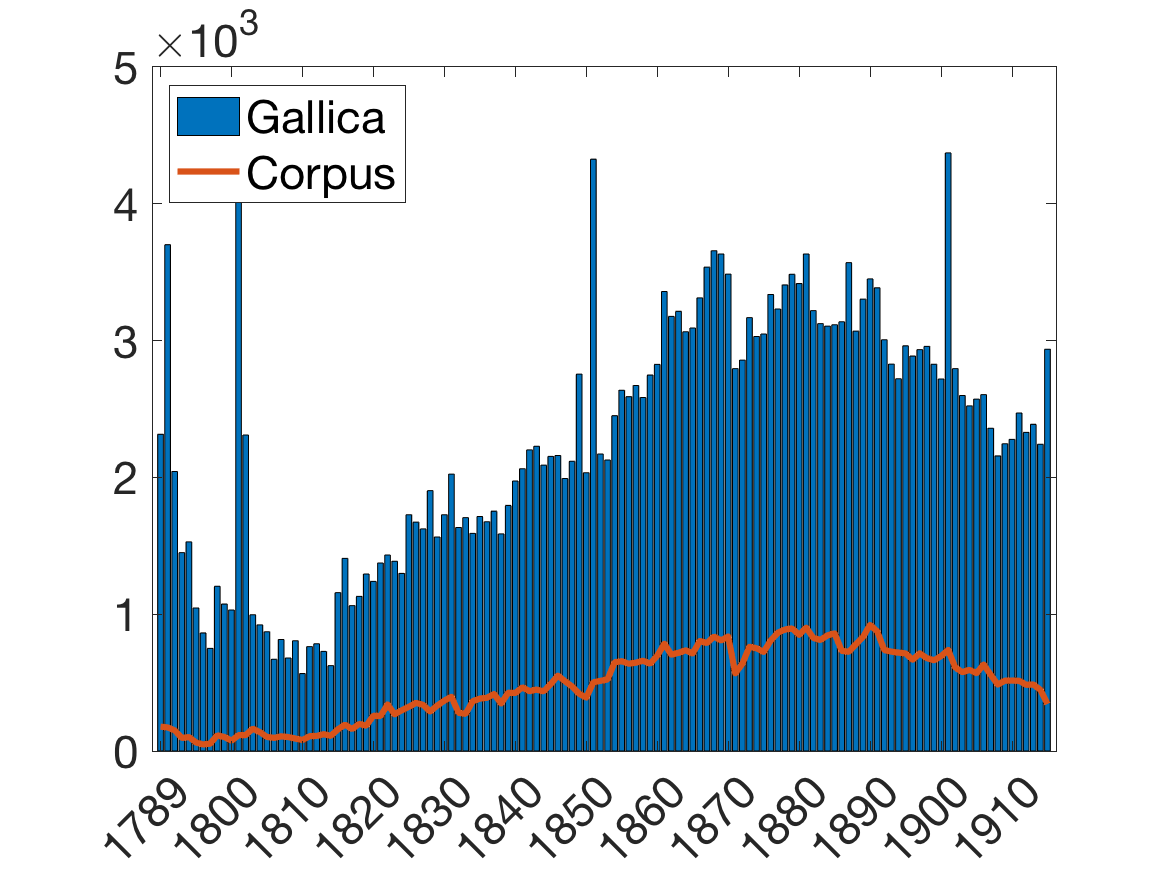}\label{fig:books_dist}}
  \hfill 
  \subfloat[Tokens distribution]{%
    \includegraphics[trim={1cm .25cm 1.5cm .15cm},clip,width=0.25\textwidth]{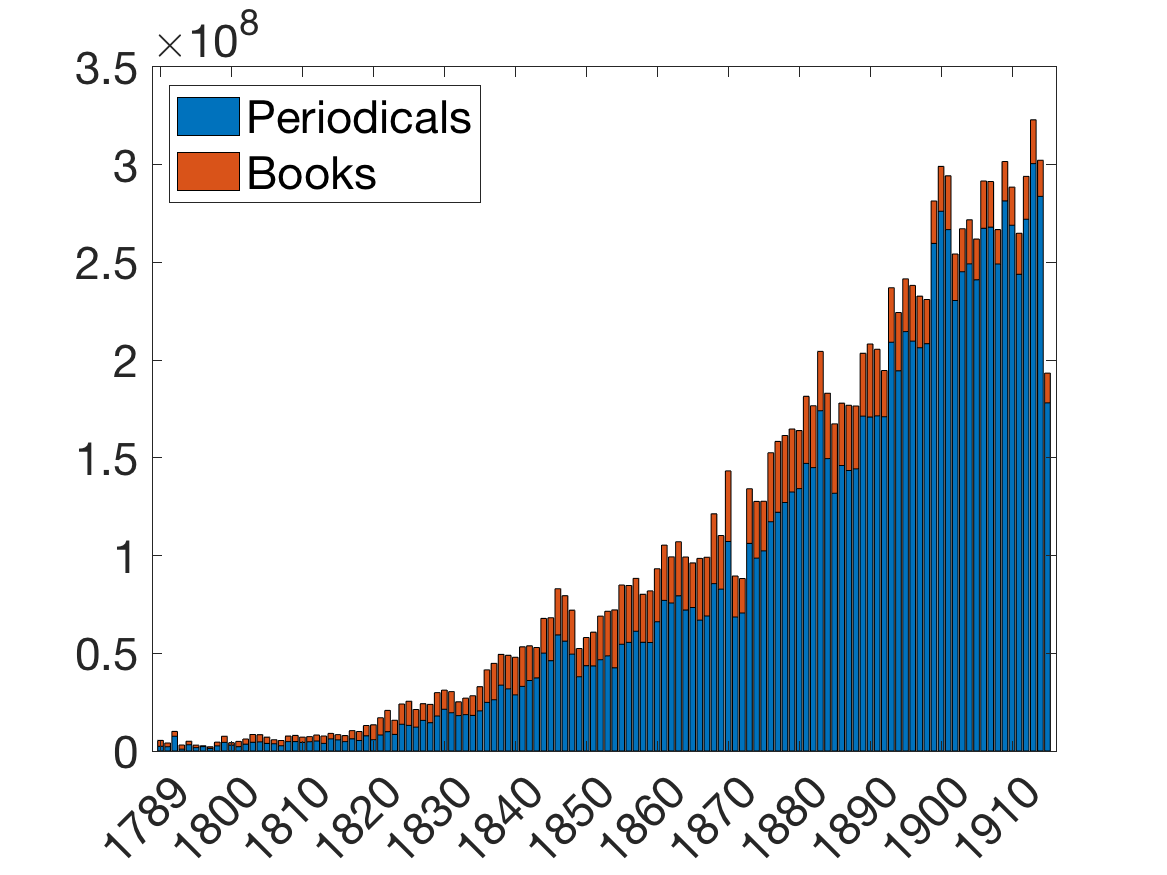}\label{fig:toks_dist}}
  \hfill 
  \subfloat[Num. of tokens in each bin]{%
    \includegraphics[trim={1cm .25cm 1.5cm .15cm},clip,width=0.25\textwidth]{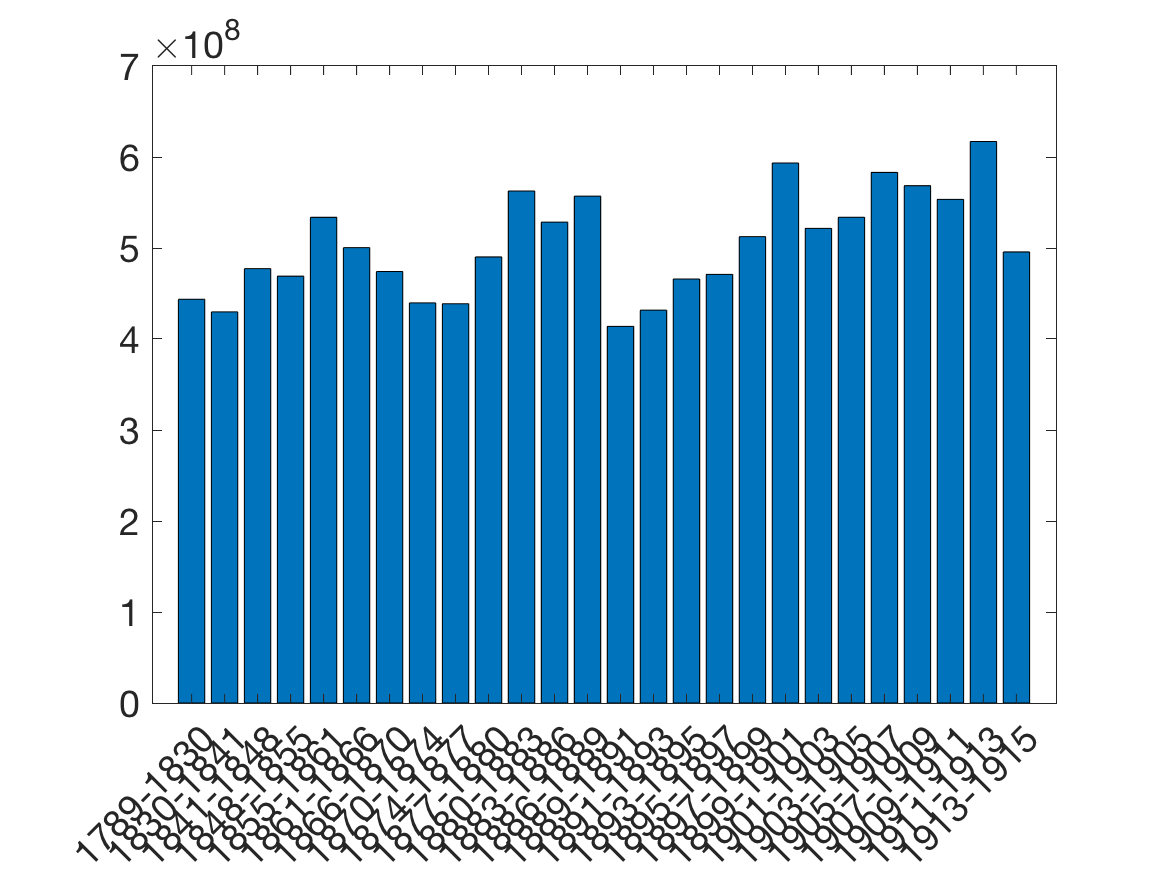}\label{fig:tok_bins_dist}}    
    \caption{Distribuition of resources in the corpus and time bins division.}\label{fig:corpora}
\end{figure*}%
Our investigation asks whether using diachronic word embeddings trained on a large corpus confirms the chronological development of antisemitic language which historians have described on a qualitative level (and if it sheds light on different, previously ignored, \emph{antisemitic moments}). We also examine the relevance of the semantic areas or streams in relation to the \emph{Jew} which we have identified based on \cite{wilson1982ideology}, and we show the trends through time of unfavourable biases towards Jews in the period considered.
\section{The Corpus and the Embeddings}
\subsection{The corpus}
The corpus\footnote{The metadata of the corpus, the embeddings and the code used for the experiments can be downloaded from https://github.com/roccotrip/antisem.} was constructed downloading from Gallica, the online library of the \emph{Bibliothèque Nationale de France}\footnote{https://gallica.bnf.fr/}, the raw text of all the resources that contain a keyword related to Jews (see appendix \ref{app:keywords} for the complete list of keywords) and have been published between 1789-1914. The research was further restricted to those resources that have an OCR quality higher that $98\%$. The resulting corpus contains $54.403$ books and $245.188$ periodicals issues. It is important to notice here that we downloaded the full text of a book or newspaper issue even if a keyword appeared only once in it.

Figures \ref{fig:per_dist} and \ref{fig:books_dist} indicate the distribution of resources per year in the periodicals and books subcorpora, respectively, together with the total number of resources in Gallica. The resources distribution per year is not homogenous in neither sub-corpora: publications increase significantly year by year. Several hypotheses can explain this proliferation of documents over time. One straightforward hypothesis can be related to increasing importance of Jews in the French public debate with the proliferation of anti-Semitic movements and newspapers such as \emph{La Croix}, \emph{La libre parole}, \emph{La Lutte antijuive} and \emph{L'Intransigeant}, just to name a few. Yet, a second hypothesis can be related to the fact that the print industry grew over time. In fact, many newspapers and publishers were founded after 1825. For example, \emph{Hachette}, the publisher with the largest number of books in our corpus ($1558$), was founded in $1826$. The newspapers \emph{Le Figaro} was founded in 1826, \emph{L'Univers} in 1833 and \emph{Le Temps} in 1861. Figure \ref{fig:per_dist} and \ref{fig:books_dist}, plotting our corpus compared to the whole Gallica one, seems to suggest that the second hypothesis is the most plausible. In fact, the quantities of resources in our corpora follow a trend similar to those observed in the whole Gallica. 
\subsection{The embeddings} Figure \ref{fig:toks_dist} shows the distribution of tokens per year distinguishing periodicals and books. The greater part of the data is from the periodicals, giving to the corpus a focus on the contemporaneity. Given this distribution it is impossible to train a model using equally sized time bins. For this reason, we decided to group the data into approximately equal bins in terms of tokens. The resulting division comprehend 26 time bins of $\approx 450$ millions tokens each (see Figure \ref{fig:tok_bins_dist}). 

For each bin we trained a word2vec skip-gram model \cite{mikolov2013distributed} using a window size of $5$ words on both sides, a word vector of $300$ dimensions and removing the words that occur less than $25$ times.
\section{Analysis}
In this section we analyse the resulting embeddings. First we study the changes in the semantic space of $4$ target words. Then we analyse the biases of the same words for $6$ different dimensions, each of which corresponds to a predetermined stream.
\begin{figure*}[!ht] 
  \subfloat[juif]{%
   \includegraphics[width=0.25\textwidth]{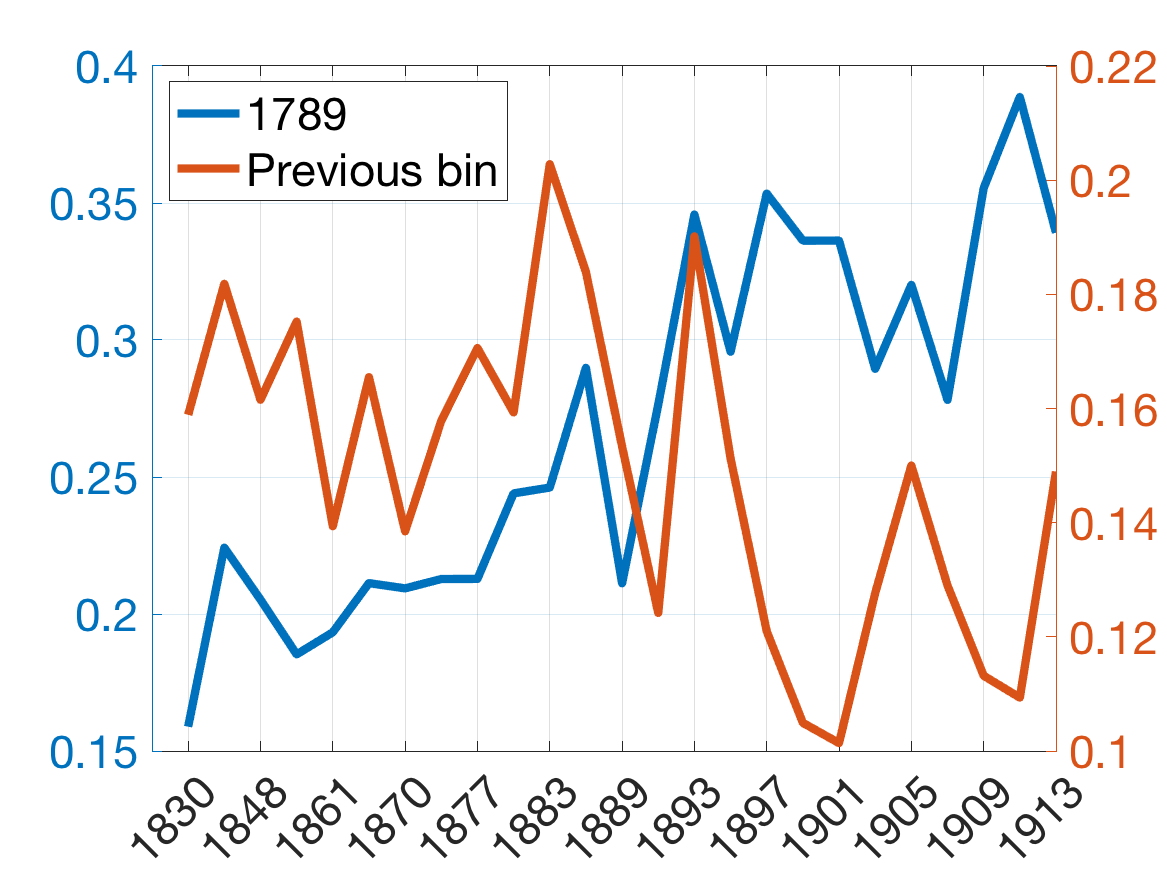}\label{fig:local_juif}}
  \hfill 
  \subfloat[juifs]{%
    \includegraphics[width=0.25\textwidth]{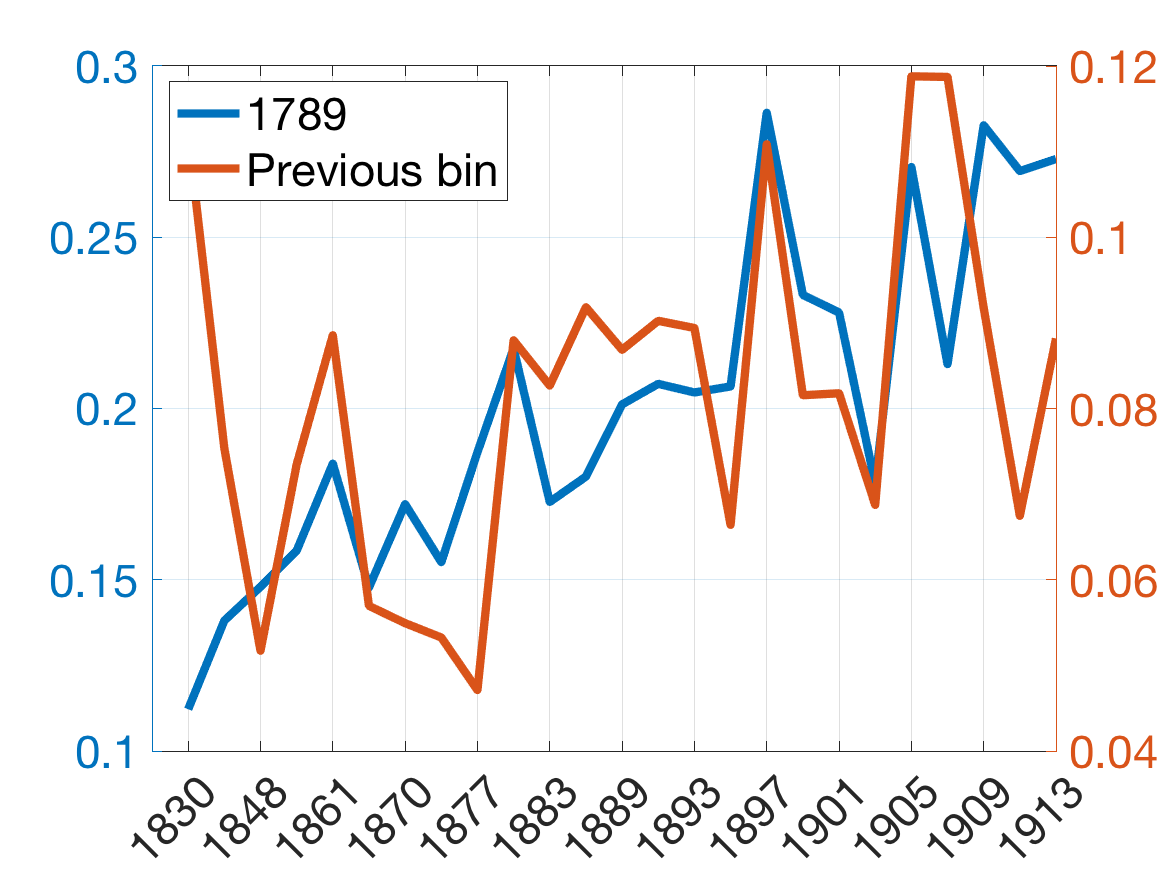}\label{fig:local_juifs}}
  \hfill 
  \subfloat[juive]{%
    \includegraphics[width=0.25\textwidth]{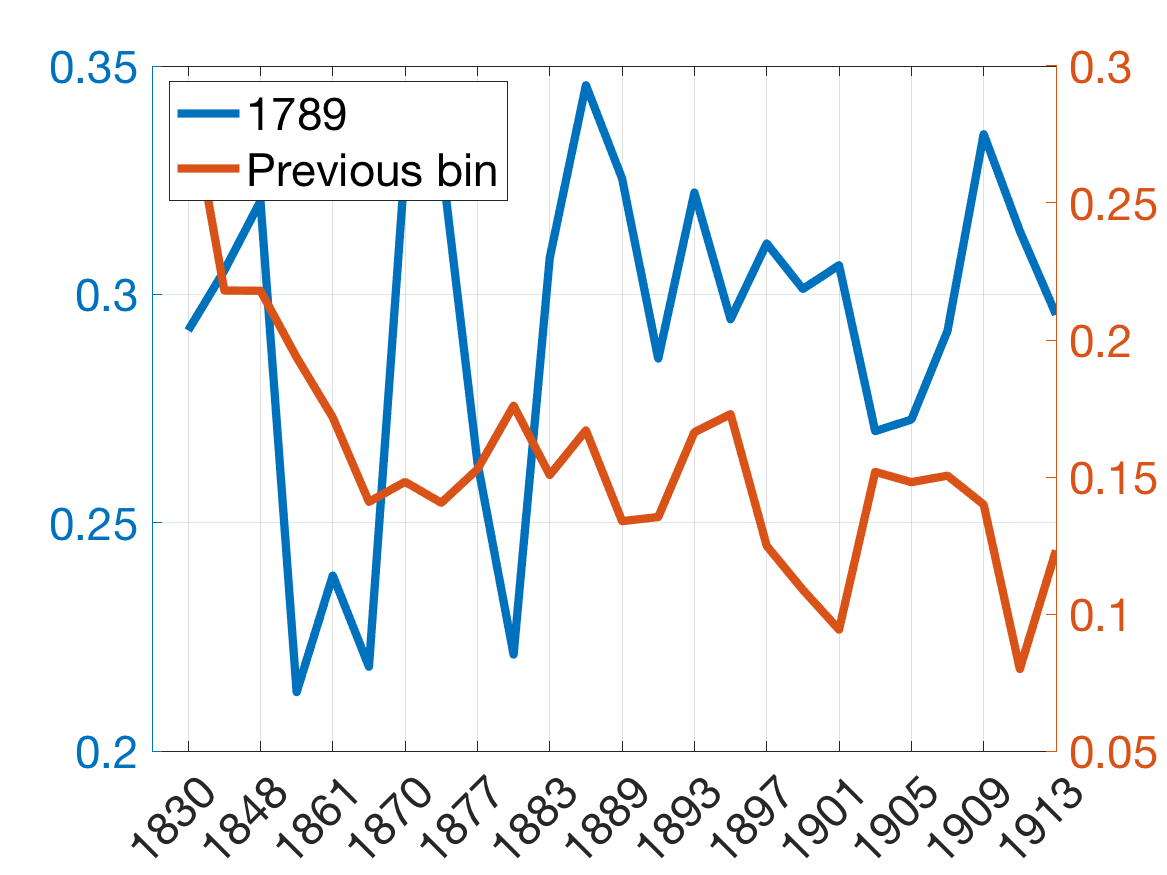}\label{fig:local_juive}}
  \hfill 
  \subfloat[juives]{%
    \includegraphics[width=0.25\textwidth]{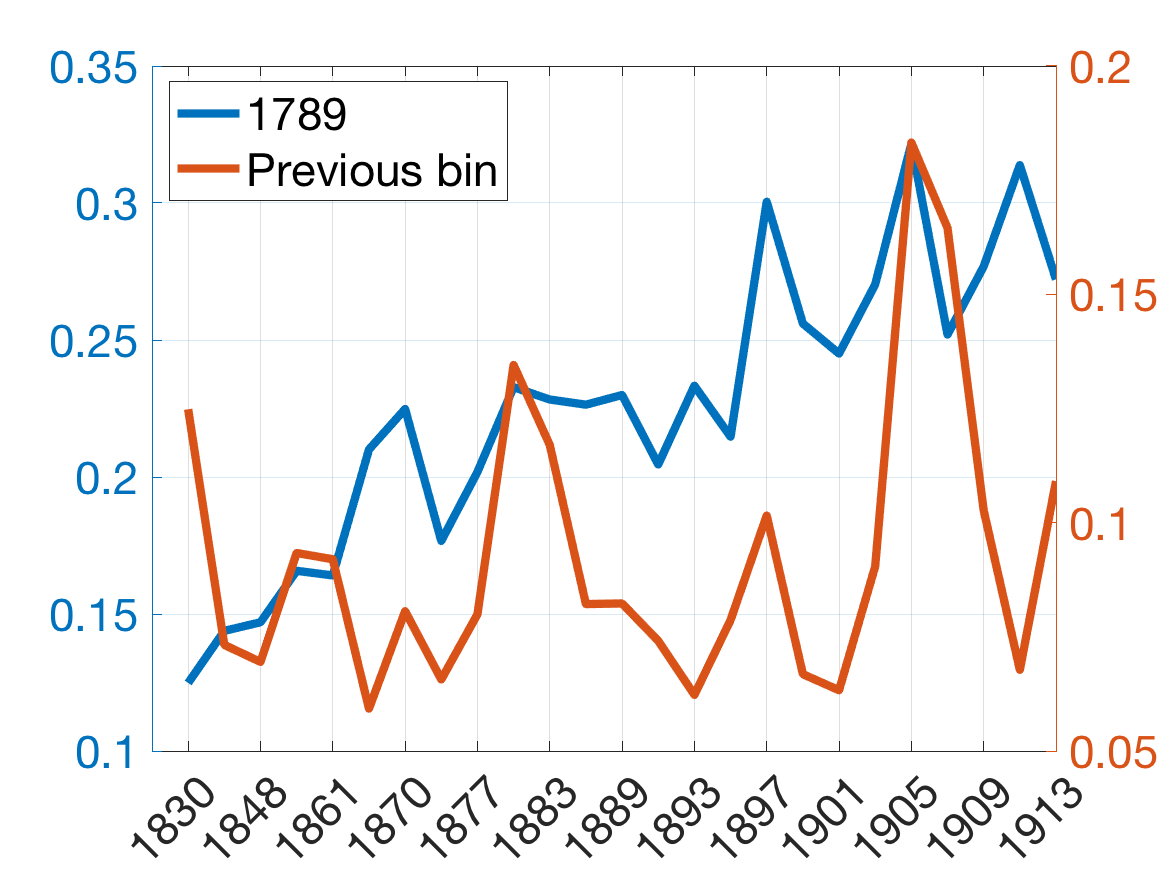}\label{fig:local_juives}}    
    \caption{Local neighborhood measure. The $y$ axes indicates the cosine distance of the second-order vector constructed for each time period compared to the $1789$ (blu line) and the preceding time period (red line).}\label{fig:local}
\end{figure*}
\begin{table*}
\resizebox{\textwidth}{!}{
\begin{tabular}{|r|l||r|l||r|l||r|l|}
\hline \multicolumn{2}{|c||}{\textbf{juif}}   & \multicolumn{2}{c||}{\textbf{juifs}} & \multicolumn{2}{c||}{\textbf{juive}} & \multicolumn{2}{c|}{\textbf{juives}}  \\
\hline \multicolumn{2}{|c||}{$\lcurvearrowdown$ \hspace{.7cm} 1841 \hspace{.7cm}$\lcurvearrowup$}   & \multicolumn{2}{c||}{$\lcurvearrowdown$ \hspace{.7cm} 1861 \hspace{.7cm}$\lcurvearrowup$} & \multicolumn{2}{c|}{ $\lcurvearrowdown$ \hspace{.7cm} 1874 \hspace{.7cm}$\lcurvearrowup$} & \multicolumn{2}{c||}{$\lcurvearrowdown$ \hspace{.7cm} 1870 \hspace{.7cm}$\lcurvearrowup$}\\
\hline
laquedem & juive & crucifient & juif & huguenots & judaïque & syriennes  & négociantes \\
mécréant & judaïque & schismatiques & israëlites & favorite & musulmane & iraniennes & samaritaines \\ 
rogatons & rabin & judaïsants & juive & opera & syrienne & musulmanes & réfugiées \\
blasphémateur & bouddhiste & fétichistes & rabbins & rigoletto & héroine & israëlites & ascètes\\ 
\hline
\multicolumn{2}{|c||}{$\lcurvearrowdown$ \hspace{.7cm} 1886 \hspace{.7cm}$\lcurvearrowup$}   & \multicolumn{2}{c||}{$\lcurvearrowdown$ \hspace{.7cm} 1870 \hspace{.7cm}$\lcurvearrowup$} & \multicolumn{2}{c||}{$\lcurvearrowdown$ \hspace{.7cm} 1886 \hspace{.7cm}$\lcurvearrowup$} & \multicolumn{2}{c|}{$\lcurvearrowdown$ \hspace{.7cm} 1880 \hspace{.7cm}$\lcurvearrowup$}  \\ \hline
ghetto & judaïque & judaïsants & juif & drumont & iranienne & israélites & épousées \\
déicides & rabin & hérétiques & synagogues & antisémitisme & apostasié & musulmanes & luthériennes \\ 
francmaçon & wanderghen & cabalistes & talmud & circoncis & lithuanienne & femmes & turques \\
aryen & anabaptiste & lucifériens & sanhédrin & théàtrale & puritaine & célébrations & dissolues\\ 
\hline \multicolumn{2}{|c||}{$\lcurvearrowdown$ \hspace{.7cm} 1893 \hspace{.7cm}$\lcurvearrowup$}   & \multicolumn{2}{c||}{$\lcurvearrowdown$ \hspace{.7cm} 1897 \hspace{.7cm}$\lcurvearrowup$} & \multicolumn{2}{c||}{$\lcurvearrowdown$ \hspace{.7cm} 1893 \hspace{.7cm}$\lcurvearrowup$} & \multicolumn{2}{c|}{$\lcurvearrowdown$ \hspace{.7cm} 1897 \hspace{.7cm}$\lcurvearrowup$}  \\ \hline
déicide & talmud & antisémites & samaritains & juiverie & synagogue & juif & dissolues \\
youtre & bouddhiste & youtres & talmud & satanisme & héroine & youtres & baptisées\\ 
francmaçon & sodomite & youpins & idolâtres & monogamique & lapidée & antijuives & prostituaient \\
youpins & anabaptiste & enjuivés & pharisiens & opprimée & persécutrice & antisémitiques & ascètes \\ 
\hline \multicolumn{2}{|c||}{$\lcurvearrowdown$ \hspace{.7cm} 1897 \hspace{.7cm}$\lcurvearrowup$}   & \multicolumn{2}{c||}{$\lcurvearrowdown$ \hspace{.7cm} 1905 \hspace{.7cm}$\lcurvearrowup$} & \multicolumn{2}{c||}{$\lcurvearrowdown$ \hspace{.7cm} 1901 \hspace{.7cm}$\lcurvearrowup$} & \multicolumn{2}{c|}{$\lcurvearrowdown$ \hspace{.7cm} 1905 \hspace{.7cm}$\lcurvearrowup$}  \\ \hline
youtre & rabin & judaïsants & synagogues & stigmatisant & dragonnade & massacrées  & courtisannes \\
sémite & usurier & hellénisants & talmud & antijuive & torturée & terrorisées & païennes \\ 
judaïsant & shylock & diaspora & pharisiens & antinationale & puritaine & diaspora & prostituaient \\
antisémite & anabaptiste & massacrant & ismaélites & dreyfusiste & anabaptiste & déportées & émigrées
 \\ \hline             
\end{tabular}}\caption{Words that have been introduced (left column $\lcurvearrowdown$) or eliminated (right column $\lcurvearrowup$) for our $4$ target words in time periods with a high local neighborhood distance, compared to 1789.}\label{tab:local}
\end{table*}%
\subsection{Local changes}
The first analysis that we conducted is the measurement of the changes in the semantic space of the words used to refer to Jews: \emph{juif} (noun/adjective, masculine, singular), \emph{juifs} (noun/adjective, masculine, plural), \emph{juive} (noun/adjective, feminine, singular) and \emph{juives} (noun/adjective, feminine, plural). For this measurement we used the local neighborhood measure proposed by \citet{hamilton2016cultural}. To compute this measure it is necessary to create a second order vector, $s$, according to equation \ref{eq:so_vector},
\begin{equation}\label{eq:so_vector}
\begin{aligned}
	s_i^{t} = \text{cos-sim}(\mathbf{w}_i^{(t)},\mathbf{w}_j^{(t)}) \forall w_j \in N_k(w_i^{(t)}) \cup \\N_k(w_i^{(t+1)}),
	\end{aligned}
\end{equation}
\noindent where $N_k(w_i^{(t)})$ represents the $k$-nearest neighbours ($k-nn$) at time ($t$) (according to cosine similarity) of a target word $w_i$ and $\mathbf{w_*}$ is the embedding corresponding to word $w_*$. Once these vectors are constructed we compute the cosine distance, $d$, among them to quantify their differences, with equation \ref{eq:local},
\begin{equation}\label{eq:local}
	d(\mathbf{s}_i^{t1},\mathbf{s}_i^{t2}) = 1 - \text{cos-sim}(\mathbf{s}_i^{t1},\mathbf{s}_i^{t2}).
\end{equation}
The results of this experiment are presented in Figure \ref{fig:local} for all the morphological variants of the word \emph{juif}, using $k=100$\footnote{We noticed that the general trend of the curves in Figure \ref{fig:local} does not change much using different values of $k$ ($10$, $25$, $50$, $100$) and that fixing $k=100$ gives a good representation of the variations over time. Increasing this value gives high fluctuations and introduces many irrelevant words.}. What emerges clearly from these figures is that there are certain periods of time in which the relation among a target word and its local neighbourhood changed consistently. What we noticed from them is that besides changes in the relative similarity among two words what changes more is the $k$-nn itself, with the introduction or elimination of specific words. 

Some of the words that were introduced (or eliminated) to (from) the $k$-nn of relevant time periods (according to local neighbourhood measure) are presented in Table \ref{tab:local}. The words in this table are ordered according to the cosine similarity with the target word. We can see an elimination of words related to the religious domain for all the target words that we used, terms like \emph{rabbin} (i.e., rabbi), \emph{talmud} (i.e. the study of the Jewish law), \emph{synagogue} and \emph{sanhédrin} (i.e., the Jewish council) are replaced by more negatively connotated words such as \emph{ghetto}, \emph{déicides}, \emph{antisémites} and \emph{antijuives}. From the few words presented in Table \ref{tab:local} one can also notice a possible rise of antisemitic prejudice (or at least of antisemitic language), with the introduction of specific words in the vocabulary specifically tailored to connote Jewish people in a derogatory way. \emph{Youtre} and \emph{youpin} are slang racist insults negatively connoting the \emph{Jew}. They appear increasingly during the period 1880-1900.

Other terms with a negative connotation that entered the semantic area of our target words are \emph{judaïsants} (i.e., judaizers), \emph{enjuivés} (i.e., strongly influenced by the Jewish spirit) and \emph{francmaçon} (i.e., freemason). These terms, as we will see in the next section, are related to the idea of a Jewish conspiracy against the world. This is a clear example of the growth of the antisemitic vocabulary.

The analysis of the word \emph{juive} is especially interesting. The word \emph{drumont} entered its space in the time period 1886-1889. It refers to Éduard Drumont, a well known antisemite who published one of the bestsellers of the antisemitism (\emph{La France juive}), in 1886,  and was the editor of an antisemitic newspaper (\emph{La libre Parole}), founded in 1892. We can also notice that in the semantic space of the word \emph{juive} there are different words related to theatre. This probably derives from literary and theatre representations of Jewish female characters, as well as references to supposed Jewish inappropriate moral and sexual behaviours. Among the theatre representations we may recall that of \emph{La Juive}, first shown in 1835, one of the most popular French operas of the 19th century, which tells the story of an impossible love affair between a Christian man and a Jewish woman. The fictional Jew, invariably seen as an outsider, provides a mirror for the phobias and obsessions of French society at a time when old Jew hatred becomes politicised, when anti-Semitism begins to permeate French ideology \cite{weinberg1983image,hallman2007opera,samuels2009inventing}.

We can also see the introduction of the word \emph{aryen} (i.e.: aryan) in 1886. This word entered the semantic space in a syntagmatic relation with the word \emph{juif} and, as we will see in the next section, the period in which it entered is characterised by a strong antisemitism characterised by an intensification of racial and socio-political stereotypes.
\begin{figure}[!ht]
\centering
\includegraphics[width=.8\columnwidth]{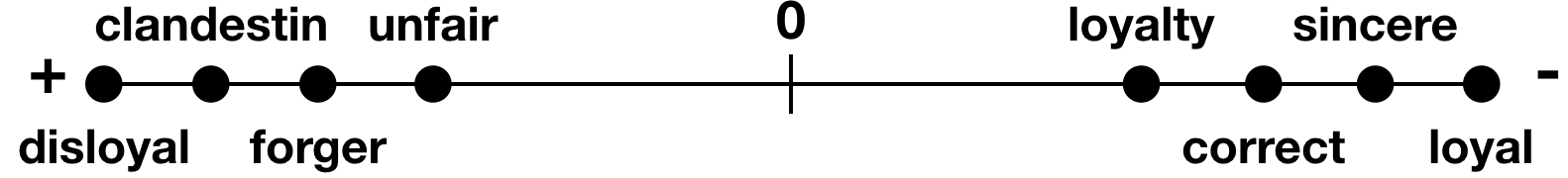}\caption{Semantic axis and projections.}\label{fig:direction}
\end{figure}
\begin{figure*}[!ht]
\centering
   \includegraphics[trim={4.5cm 1.1cm 4.3cm .0cm},clip,width=.95\textwidth]{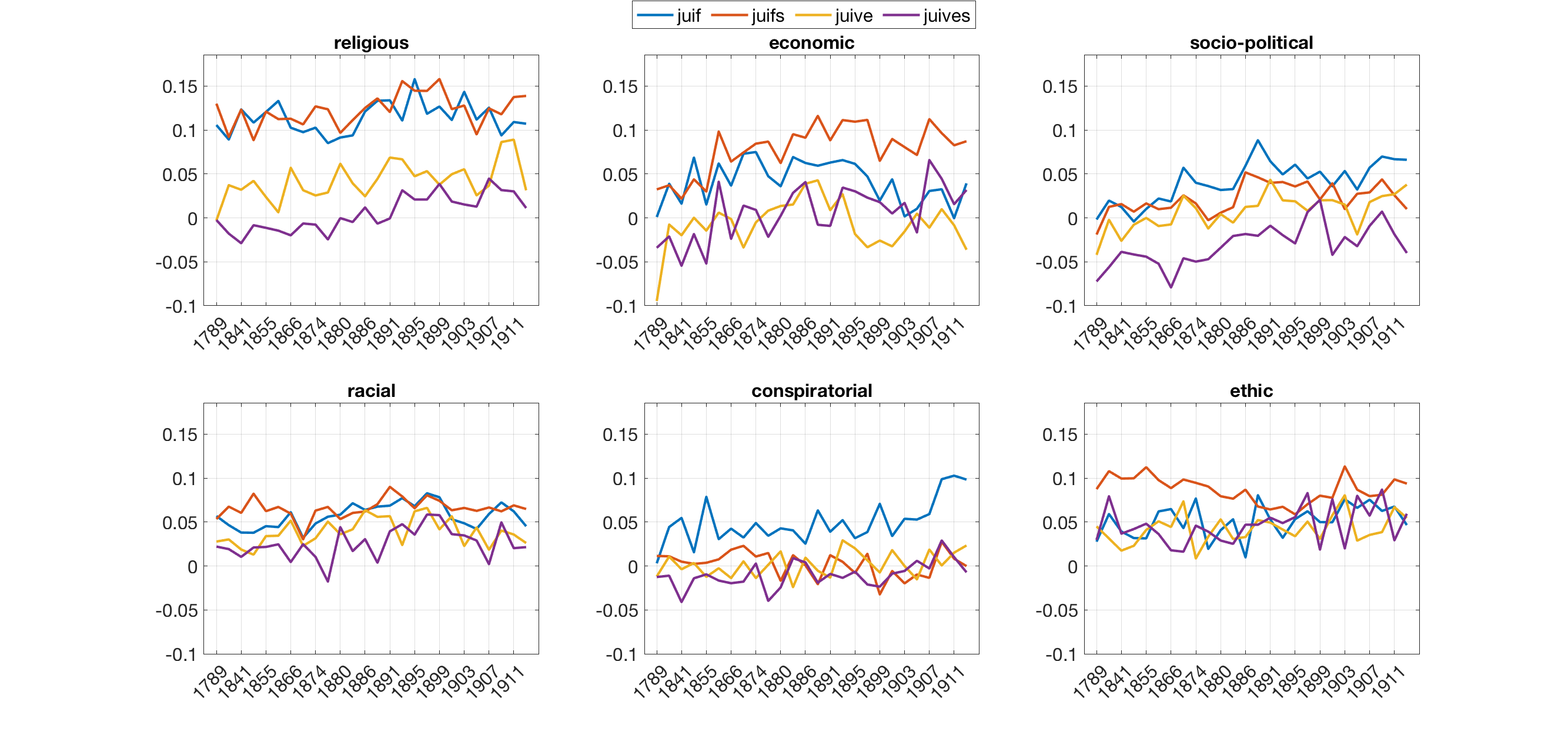}\caption{Projections of our 4 target words to the 6 semantic axes. Positive values indicates the adverse bias.}\label{fig:bias_streams}
\end{figure*}
\subsection{Embedding projections}
\subsubsection{The streams}
To quantify biases in word embeddings semantic spaces it is common to project a specific word vector on a semantic axis \cite{bolukbasi2016man,caliskan2017semantics}. The semantic axis can be computed as $\mathbf{g}=\mathbf{w}_i - \mathbf{w}_j$ and its projection as the dot product $\hat{b} = \mathbf{w} \cdot \mathbf{g}$, assuming that the vectors are normalised, the projection is equal to the cosine similarity. The higher the values of the projection, the more biased the word is toward that direction.

In previous literature \cite{bolukbasi2016man} the gender direction (e.g., $\vec{he} - \vec{she}$) was used to project words related to occupations in order to quantify if these words embed information about gender. In this work we do not want to project words only according to a single direction but we want to analyse different adverse and (or) favourable biases, comparing them over time. For this reason, we defined six different semantic axes, that correspond to six antisemitic streams ($S$) \cite{wilson1982ideology}.

For each stream, $s \in S$, we identified a set of $n$ antonyms pairs, $z_s=\{(a_1^{neg},a_1^{pos}), ..., (a_n^{neg},a_n^{pos})\}$ to construct the bias subspace in the embedding. To avoid selection biases we selected the antonyms pairs starting from a positive seed word, that is highly representative for the stream, and used a knowledge base to collect its synonyms and the corresponding antonyms (see appendix \ref{app:proj} for the complete list of antonyms used). We noticed that computing the PCA of each subspaces the corresponding explained variance is concentrated on the first component and that it is stable over time. For example, the first component of the racial stream has an explained variance of $0.34$ (mean) with a standard deviation of $0.012$.

The six different semantic areas, which may correspond to related antisemitic discourses are:
\begin{enumerate}
\item religious: antisemitism based on theological doctrines or narratives, and on religious prejudices and accusations. The seed word is \emph{believer} (\emph{unbeliever});
\item economic: antisemitism based on a supposed Jewish role in the economy or on stereotypes concerning Jews' economic behaviours. The seed word is \emph{generosity} (\emph{greed});
\item socio-political: antisemitism based on malevolent, e.g. anti-national, political behaviours or on supposedly threatening Jewish actions. The seed word is \emph{honor} (\emph{shame});
\item racial: antisemitism based on the definition of Jews as a race, considered inferior. The seed word is \emph{pure} (\emph{impure});
\item conspiratorial: antisemitism based on conspiracy theories. The seed word is \emph{loyal} (\emph{disloyal});
\item ethic: antisemitism based on Jewish supposed unethical or perverse morals or behaviours. The seed word is \emph{moral} (\emph{immoral});
\end{enumerate}
\noindent To quantify the biases for all the time we computed the mean bias, $b$, for each stream as the arithmetic mean of the individual biases, $\hat{b}$ on each axis, according to equation \ref{eq:mean_bias}:
\begin{equation}\label{eq:mean_bias}
	b(w_i, s) = \frac{1}{n}\sum_{j=1}^{n}{\mathbf{w}_i\cdot (\mathbf{w}_{a_{j}^{neg}} - \mathbf{w}_{a_{j}^{pos}}}),
\end{equation}
\noindent where $n$ is the number of antonyms pairs in stream $s$, Given the ordering of the antonyms in the computation of the bias axis ($\mathbf{g}=\mathbf{w}_{a_{j}^{neg}} - \mathbf{w}_{a_{j}^{pos}}$) we define an adverse bias when $b$ is positive and a favourable bias when $b$ is negative. 

An example of semantic axis constructed with the pair \emph{disloyal} as negative word and \emph{loyal} as positive, is presented in Figure \ref{fig:direction} ($\vec{disloyal} - \vec{disloyal}$). From this figure we can see that words that have a high projection value are words very similar to the negative word, on the other hand, words with a low projection are very similar to the word on the other side. The projection tells us if a word is closer to one extreme or the other. Unbiased words should have a projection close to $0$.%
%
\subsubsection{Biases related to \emph{Jews}}
The results of this experiment are presented in Figure \ref{fig:bias_streams}. The adverse bias is always high for the words \emph{juif}, \emph{juifs} and \emph{juive}. For the word \emph{juives} only on few cases it is negative. Adverse and favourable biases are measured with positive and negative measures respectively.

Our analysis confirms the chronological development of \emph{antisemitic moments} identified by historians, with a steady increase of adverse bias starting in the 1880s, before the Dreyfus affair. We also notice an unexpected peak in adverse bias between 1855 and 1866, in connection with the French Second Empire (1851-1870). The semantic areas or streams in relation to the Jew identified on the basis of \cite{wilson1982ideology} seem relevant for the description of adverse bias in \emph{antisemitic moments}. 
\begin{figure*}[!ht]
\centering
   \includegraphics[trim={4.9cm .45cm 4.3cm .35cm},clip,width=1\textwidth]{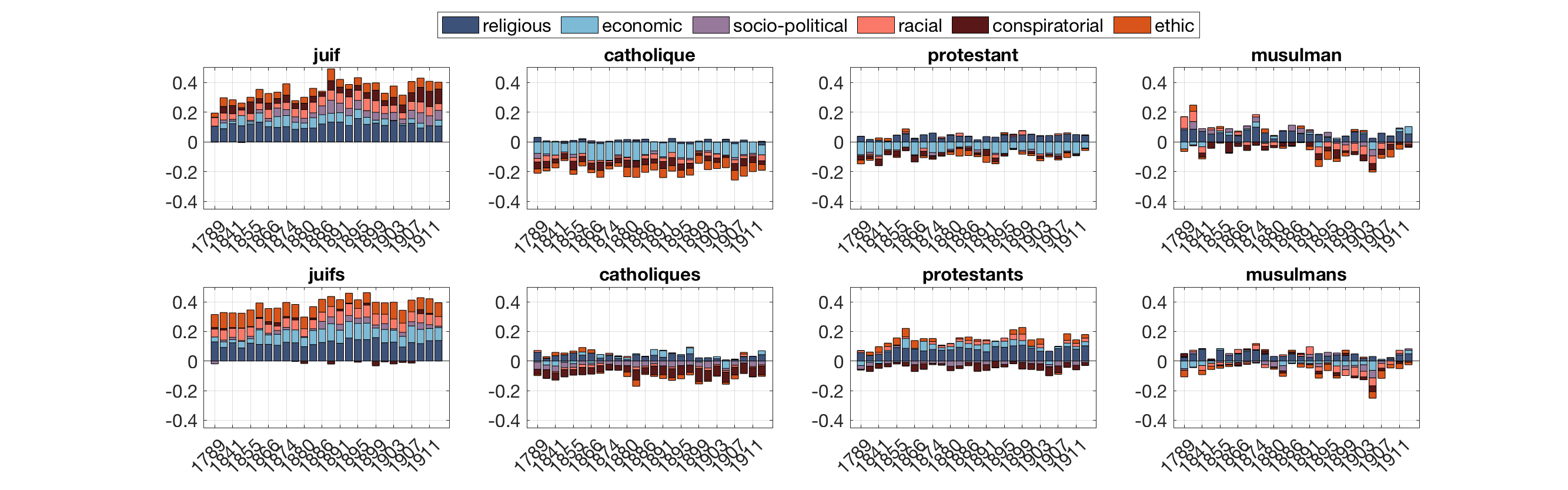}\caption{Cumulative bias projections compared to different religious groups.}\label{fig:bias_streams_cum}
\end{figure*}
\begin{figure*}[!ht]
\centering
   \includegraphics[trim={4.9cm .45cm 4.2cm .35cm},clip,width=1\textwidth]{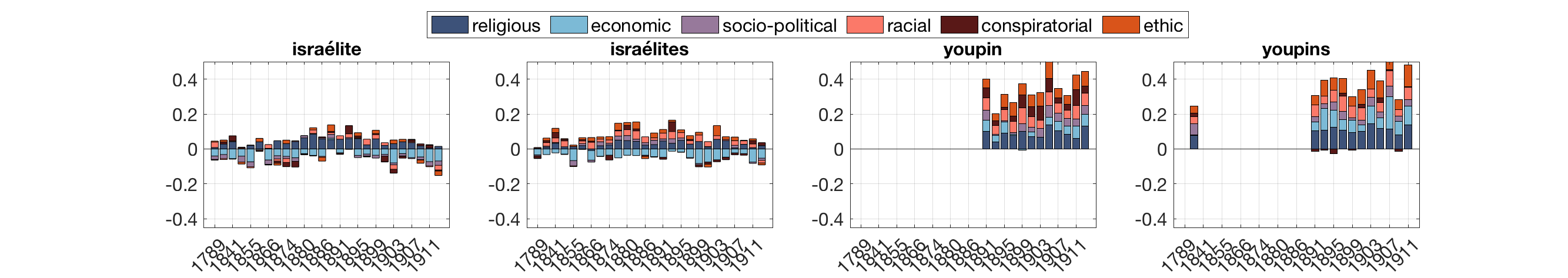}\caption{Cumulative bias projections for other words used to refer to Jews.}\label{fig:bias_streams_cum2}
\end{figure*}
The highest adverse bias characterises the religious semantic area, followed by the economic and ethic areas. The religious adverse bias shows a peak starting in 1855, after the establishment of Napoleon III's Second Empire, a time of renewed allegiance to the Catholic Church and in 1895 at the beginning of the Dreyfus affair. Also the economic adverse bias shows a peak starting in 1855, perhaps because of the increase of economic discourse on Jews following the publication of Toussenel's \emph{Les Juifs rois de l'époque}, and again coinciding with the establishment of the Second Empire. Another peak comes with the Dreyfus affair. The ethic adverse bias peaks in the period 1830-1855, diminishes afterward and peaks again toward the end of the Dreyfus affair.

Racial, conspiratorial and especially sociopolitical semantic areas show a steady adverse bias and an increase mostly after 1886, i.e. after the publication of Drumont's La France juive (1886). The conspiratorial adverse bias also peaks – like the religious, the economic and the ethic adverse bias – in 1855.

The singular \emph{juif} prevails in the conspiratorial and socio-political semantic areas, which seem to entail general statements about \emph{the Jew}. This tendency has been noticed by historians as typical of modern antisemitism and has been called \emph{singularisation} \cite{miccoli2003antiebraismo}. This underlines that there are features \emph{common to all [Jews], because in all and every one there emerges something which constitutes a common and exclusive feature} of the Jew as \emph{the enemy to be defeated} \cite{miccoli2003antiebraismo}. On the other hand, the plural \emph{juifs} prevails in the economic and ethic areas, as implying collective behaviours of Jews.

Racial, sociopolitical and conspiratorial semantic areas show a steady adverse bias and increase especially after 1886. As the racist vision of the Jew increases, it is turned increasingly into a political vision, and it is also nourished by a conspirationist worldview, which will culminate in the Dreyfus affair.
\subsubsection{Comparative biases concerning different religious groups}
The results of this experiment are presented in Figure \ref{fig:bias_streams_cum}. They show a comparison with three different religious groups: Catholic, (\emph{catholique}), Protestant (\emph{protestante}) and  Muslim (\emph{musulman}). The plots sum positive and negative biases to give a general picture of the biases at each time step.

\emph{Juif} and \emph{catholic} have a completely opposite bias: exclusively adverse in the first case, entirely favorable in the latter. Confronting \emph{juif} and \emph{protestant} we notice a similar bias, adverse in the first case, favorable in the latter. But the favorable bias of Protestant is much more reduced than that of \emph{catholic}. Confronting \emph{juifs} and \emph{protestantes}, both show an adverse bias (lower in the case of Protestants). The adverse bias concerns \emph{protestantes} especially in relation to the religious domain. \emph{Musulman} and \emph{musulmans} also show an adverse bias concentrated in the religious sphere.
\begin{figure*}[!ht]
\centering
  \subfloat[]{%
   \includegraphics[trim={1cm .25cm 1.5cm .15cm},clip,width=0.25\textwidth]{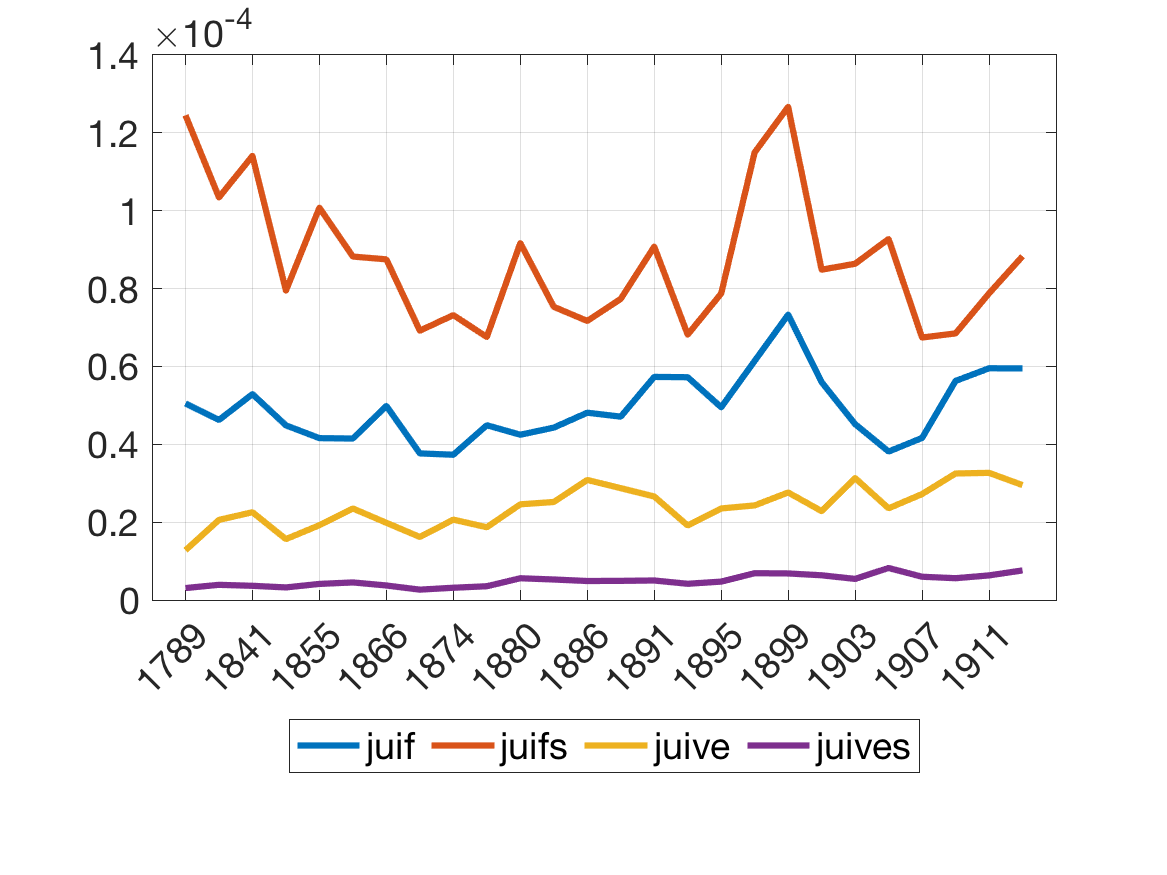}\label{fig:frequencies1}}
  \hfill 
  \subfloat[]{%
    \includegraphics[trim={1cm .25cm 1.5cm .15cm},clip,width=0.25\textwidth]{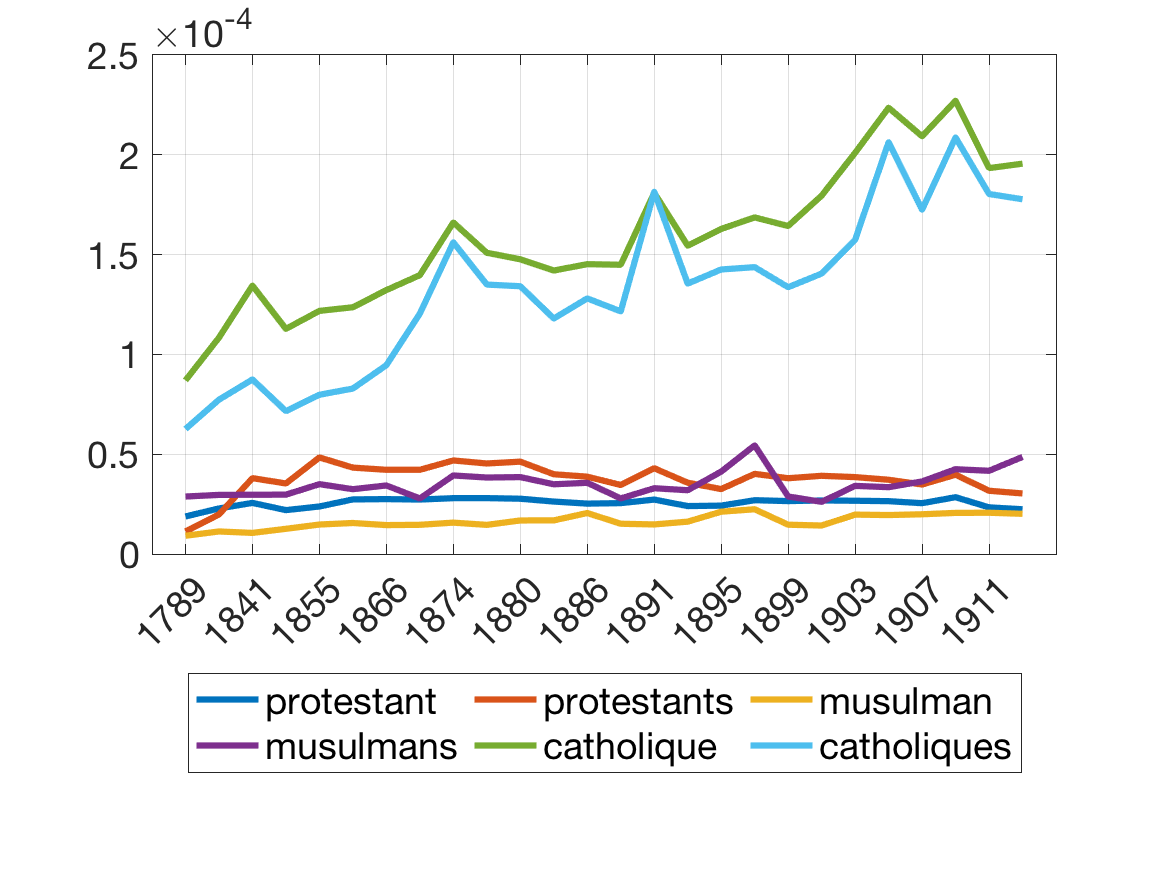}\label{fig:frequencies2}}
  \hfill 
  \subfloat[]{%
    \includegraphics[trim={1cm .25cm 1.5cm .15cm},clip,width=0.25\textwidth]{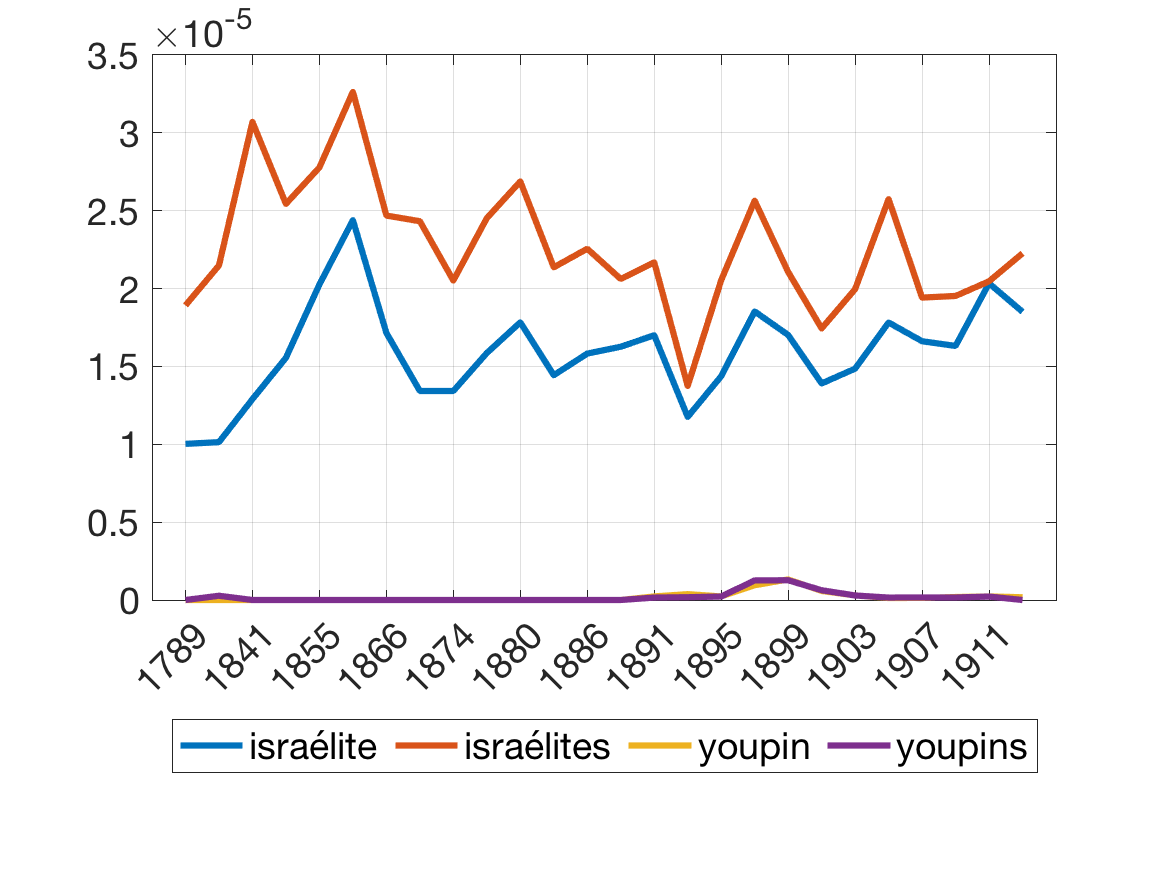}\label{fig:frequencies3}}\caption{Target words frequency.}  
\end{figure*}%
If we look at racial stream, this grows for \emph{juif(s)} reference to \emph{protestants} is absent; while there is an occasional emergence of \emph{musulman}, with an adverse bias between 1789 and 1840, when questions of citizenship are being defined (France conquers Egypt in 1798 and in 1834 Algeria is annexed to France; in 1870 the Crémieux Decree granted French citizenship to Algerian Jews but not to Muslims), and a favourable bias in 1891-95 (in 1890 a bill is proposed for the granting of French citizenship to Algerian muslims, see Weill, 2005). The last increase is probably also connected with the availability of a larger quantity of digitised North-African press in the corpus.
\subsubsection{Comparative bias concerning different words used to refer to Jews}
The results of this experiment are presented in Figure \ref{fig:bias_streams_cum2}. \emph{Israelite} and \emph{israelites} do not show a particular bias as the terms are often used euphemistically (including by Jews themselves), i.e. preferred to the more direct and connotated \emph{juif} and \emph{juifs}. These terms refers to the cultural assimilation and social integration of Jews into French society, as described by \citet{honore1981vocabulaire}.

The slang and derogatory \emph{youpin} spread starting around 1886; its shows an exclusively adverse bias and a trend similar to \emph{juif}, as if the terms \emph{juif} and \emph{youpin} were interchangeable.
\subsection{Target words frequency}
Even if the corpus has been constructed selecting documents containing words related to the Jewish question, we noticed that the frequencies of words related to other religious groups is higher for \emph{catholique} and \emph{catholiques} and slightly lower for the words \emph{protestant}, \emph{protestantes}, \emph{musulman} and \emph{musulmanes}. The frequencies of all the target wordsare reported in Figure \ref{fig:frequencies1}, \ref{fig:frequencies2} and \ref{fig:frequencies3}.
\section{Conclusions}
References to Jews increase throughout the 19th century, as Jews were integrated within French society and these references appear to be mostly associated with an adverse bias in all semantic areas. The adverse bias grows starting in the mid-1880s, i.e. in the second half of the Third Republic, when the rise of anticlericalism and socialism was associated with Jews by the conservative and catholic public opinion. Around this time the publication of Drumont's \emph{La France juive} provokes an adverse bias towards Jews clearly associated to antisemitic discourse in all semantic areas, which prepares the outburst of the Dreyfus affair, and it remains steady during and after the affair. 

The highest adverse bias characterises the religious semantic area, followed by the economic and ethic spheres. The conspiratorial and sociopolitical areas show an adverse bias more often associated with the singular \emph{juif}, as if they provoked categorical statements.  Adverse bias in the economic and ethic areas is expressed through the plural \emph{juifs} as describing collective behaviours.

The confrontation between \emph{juif} and \emph{catholic} shows an entirely adverse bias in the first case and an entirely favorable bias in the latter case. The adverse bias towards other minorities, i.e. Protestants and Muslims concerns the religious semantic area. No bias concerning protestant emerges in the racial semantic area, while a negative and positive bias emerge in relation to Muslims at times when the question of French citizenship is being defined.

As one evaluates the presence of the word \emph{juif}, and the semantic areas surrounding it, one should also consider that these may emerge in texts which are not antisemitic per se, but still contribute to the spread of images of Jews, with specific biases. We refer here especially to literary texts.

We suggest that the adverse bias in various semantic areas may be associated with antisemitic discourses, but this association should be further explored though an examination of the historical context (for example that of \emph{antisemitic moments}) or an analysis of the textual sources which spread the words associated with \emph{the Jew}.
\section*{Acknowledgments}
\begin{wrapfigure}{i}{0.03\columnwidth}
\raisebox{0pt}[\dimexpr\height-0.9\baselineskip\relax]{\includegraphics[width=0.09\columnwidth]{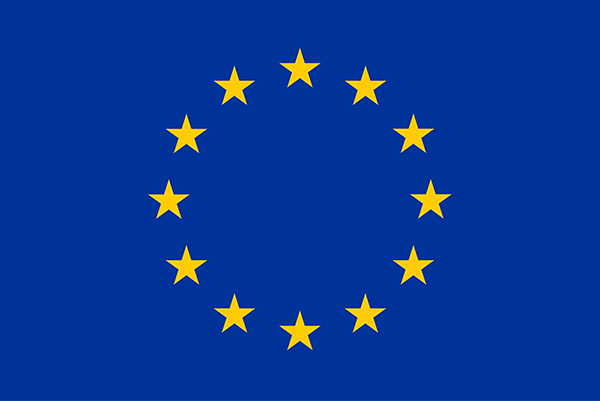}}%
\end{wrapfigure}%
The authors of this work have received funding from the European Union's Horizon 2020 research and innovation programme under grant agreement No 732942. The experiments have been run on the SCSCF cluster of Ca' Foscari University.
\bibliography{naaclhlt2019}
\bibliographystyle{acl_natbib}
\appendix
\section{Keywords}\label{app:keywords}
\begin{itemize}
        \item Juif (i.e: Jew - masculine, singular)
        \item Juive (i.e: Jew - feminine, singular)
        \item Judaisme (i.e: Judaism)
        \item Isra\"{e}lite (i.e: Israelite)
        \item Isra\"{e}l (i.e: Israel)
        \item Isra\"{e}litisme (i.e: Israelitism)
        \item Mosa\"{i}sme (i.e: religions referred to the message of Moses)
        \item Talmud (i.e: Talmud)
        \item Judas (i.e: Judass)
        \item Moloch (i.e: the biblical name of a Canaanite god associated with child sacrifice)
        \item Ahasverus (i.e: a mythical immortal man whose legend began to spread in Europe in the 13th century. The original legend concerns a Jew who taunted Jesus on the way to the Crucifixion and was then cursed to walk the earth until the Second Coming.)
\end{itemize}
\section{Bias axes}\label{app:proj}
The list of antonyms used to compute the bias axes. Note that the translation of the antonyms pairs is provided only for the singular. We used a public resource (http://www.synonyms-fr.com) to collect antonyms relations.
\paragraph{Religious} angel, devil; sacred, profane; pious, atheist; pious, pagan; pious, idolater; pious, impious; sacred, cursed; venerable, abject; faithful, unfaithful; believer, unbeliever; religious, irreligious; dedicated, atheist.
\paragraph{Economic} give, appropriate; generosity, greed; generous, greedy; generous, miserly; generous, stingy.
\paragraph{Socio-political} prodigal, greedy; honest, rabble; honor, shame; friendly, hostile; loyal, deceitful; socialist, capitalist; friend, enemy; ally, antagonist; conservative, progressive.
\paragraph{Racial} normal, strange; superiority, inferiority; equality, inequality; pleasant, unpleasant; benign, wicked; worthy, infamous; sympathy, hate; accepted, refused, better, worse; national, foreign; pure, impure; upper, lower; pure, filthy; clean, dirty.
\paragraph{Conspiratorial} loyal; spy; honesty, treason; loyal, disloyal; clear, mysterious; obvious, occult; sincere, deceitful; sincere, unfair; benefactor, criminal; clear, secret; friendly, threatening; clear, dark.
\paragraph{Ethic} chastity, lust; modest, intriguing; decent, indecent; virtuous, lascivious; faithful, unfaithful; moral, immoral; honest, dishonest; chaste, depraved; chaste, fleshly; pure, degenerate.
\end{document}